\documentclass[11pt,a4paper]{article}
\usepackage[hyperref]{emnlp2020}
\usepackage{times}
\usepackage{latexsym}

\usepackage{microtype}
\usepackage{amsmath,amssymb}
\usepackage{comment}

\usepackage{xspace}
\usepackage{booktabs}
\usepackage{multicol}
\usepackage{amsmath}
\usepackage{multirow}
\usepackage{graphicx}
\usepackage{enumitem}
\usepackage{mdwlist}

\newcommand{\sref}[1]{\S\ref{#1}}
\newcommand{\fref}[1]{Figure~\ref{#1}}

\newcommand{\myparagraph}[1]{\paragraph{#1}}

\aclfinalcopy 
\def\aclpaperid{3352} 

\title{Quantifying Intimacy in Language}

\author{Jiaxin Pei \\
  School of Information \\
  University of Michigan \\
  \texttt{pedropei@umich.edu} \\\And
  David Jurgens \\
  School of Information \\
  University of Michigan \\
  \texttt{jurgens@umich.edu} \\}

\date{}

\begin{document}
\maketitle
\begin{abstract}
 Intimacy is a fundamental aspect of how we relate to others in social settings. Language encodes the social information of intimacy through both topics and other more subtle cues (such as linguistic hedging and swearing). Here, we introduce a new computational framework for studying expressions of the intimacy in language with an accompanying dataset and deep learning model for accurately predicting the intimacy level of questions (Pearson's $r$=0.87). Through analyzing a dataset of 80.5M questions across social media, books, and films, we show that individuals employ interpersonal pragmatic moves in their language to align their intimacy with social settings. Then, in three studies, we further demonstrate how individuals modulate their intimacy to match social norms around gender, social distance, and audience, each validating key findings from studies in social psychology. Our work demonstrates that intimacy is a pervasive and impactful social dimension of language.

\end{abstract}

\section{Introduction}

    

Intimacy is a vital ingredient in the hierarchy of human needs \cite{maslow1981motivation,erikson1993childhood,sullivan2013interpersonal}, playing key roles in development \cite{harlow1959affectional} and well-being \cite{sneed2012relationship}.
Language provides multiple means of conveying intimacy in a social context as individuals make decisions on the topic of conversation, phrasing, and markers relating the speaker to the world. 
Much like how social status and power are constructed and expressed, individuals negotiate intimacy in language to fulfill fundamental and strategic needs, while respecting social norms about the appropriate intimacy \cite{chaikin1974variables,korobov2006intimacy}.
In this paper, we aim to quantify the intimacy expressed in language and demonstrate how this intimacy is constructed and employed across diverse social settings.

While sociolinguistics and social psychology have long pointed to how people shape their language to convey social information \cite{labov1972language,brown1978universals,clark1980polite,weber2008handbook,locher2010interpersonal}, only recently, have computational models focused on making this information explicit \cite{choi2012hedge,danescu2013computational,bak2014self}. In particular, works on social status and power have shown how individuals use lexical cues and linguistic strategies like accommodation to express their perceived status in relation to others \cite{danescu2013computational,prabhakaran2014gender}.
Much like status in society, intimacy is a natural concept describing how an individual relates to their audience in their perceived interdependence, warmth, and willingness to personally share  \cite{perlman1987development}.
Our work provides the first model of intimacy in language and tests its implications.

In this paper, we examine the intimacy of questions. As requests for information, questions provide a natural mechanism for studying how people shape the intimacy of their questions in response to the social context \cite{clark1980polite,jordan1990acquiring}. Questions serve a fundamental role in dialogues for interpersonal exchange \cite{athanasiadou1991discourse}, and increasingly intimate questions are known to foster deep social ties \cite{aron1997experimental,kashdan2011curiosity}. 
Our work starts from a new dataset of 2,397 questions rated for intimacy using Best-Worst-Scaling \cite{louviere2015best,bws-naacl2016}. We use this dataset to train a deep learning model that obtains high correlation with human judgment on both in-domain and out-domain settings. Using this model, we rate 80,484,458 questions across Reddit, Twitter, literature, and film.

Through this massive dataset, we demonstrate how individuals actively construct their social context through linguistic choices that signal the acceptability of a question's intimacy.
Given the role of intimacy in social interactions, we examine the social perspective of intimacy in conversations in three settings. First, we show that the intimacy level of language reflects cultural norms of masculinity and femininity, which our study shows persist across real and imagined settings---and even across other gender's perceptions of the norms. Second, even online, individuals reserve their most intimate questions for close friends and strangers, mirroring offline observations, where the risk of social capital loss is greatest for acquaintances. Finally, online communication affords a new communication mechanism---complete anonymity---for communicating, which we demonstrate is used as a strategy to ask more intimate questions.
Both the model and datasets used in this paper are released at \url{https://blablablab.si.umich.edu/projects/intimacy/}.

\section{Theories of Intimacy}

As a natural concept in social settings \cite{helgeson1987prototypes}, intimacy has been widely explored in interpersonal, behavioral, and social domains \cite{prager1995psychology,weber2008handbook,locher2010interpersonal}.  
Studies of intimacy in communication have primarily focused on the exchange of personal information as a measure of intimacy \cite{miller1982assessment,descutner1991development}. 
Yet, intimacy in language is revealed more than just through disclosure, as individuals select topics, phrasings, and styles to indicate their intimacy with another within the social context.
Here, we study the role of intimacy in language and argue that intimacy is a natural component of language similar to politeness \cite{brown1978universals}. 
Following, we discuss the theoretical background of intimacy in social psychology and linguistics.

\myparagraph{The Concept of Intimacy}
The concept of intimacy has long been proposed by psychologists with various definitions. While intimacy generally refers to the closeness and interdependence of partners, the extent of self-disclosure, and the warmth or affection experienced within the relationship \citep{perlman1987development}, the concept of intimacy is not restricted to the closeness or interactions between people in intimate relationships, as even people who are not in intimate relationships can have intimate interactions in a certain space and time \cite{wynne1986quest}. Therefore, a widely-accepted conceptualization of intimacy is to distinguish between intimate interactions and intimate relationships \cite{hinde1981bases}. The former refers to dyadic communicative exchanges, while the latter is the history and future expectations of intimate contact over time \cite{prager1995psychology}. Intimate interactions and intimate relationships are interrelated in the following way: while intimate interactions are necessary in the formation of intimate relationships and are regular within them \cite{prager1995psychology}, intimate relationships also build expectations for the depth and types of interactions between people \cite{chelune1979self}. 

In this work, we focus on the language aspect of intimate interactions as dialogue is one of the core forms of intimate interactions \cite{hinde1981bases} and certain ways of communications can facilitate the experience of interpersonal closeness \cite{aron1997experimental}.

\myparagraph{Intimacy in Interpersonal Communication}
Language is one of the core aspects of intimate interactions \cite{hinde1981bases} and multiple branches of linguistics have studied aspects of communications in social relationships which relate to intimacy, including interactional sociolinguistics \cite{gumperz201511}, conversational analysis \cite{hutchby2008conversation}, and discourse analysis \cite{johnstone2018discourse}.

Most notably, works in interpersonal pragmatics have analyzed the relational aspect of interactions between people that both affect and are affected by their understandings of culture, society, and their own and others' interpretations \cite{locher2010interpersonal}.
One closely-related branch of interpersonal pragmatics is the study of politeness \cite{brown1978universals}, which demonstrates how people change their expressions to construct politeness with respect to different social settings to signal relative power. Analogously, intimacy in language can also be constructed with intentional pragmatic choices to signal the perceived intimacy between speakers. However, while psychologists have deeply explored people's behavior of self-disclosure \cite{cozby1973self} as one of the major components of verbal intimacy \cite{fitzpatrick1987marriage}, intimacy in language is not just conveyed by the degree of self-disclosure. The style of the language can indicate the intimacy of the speaker \cite{bell1984language}, e.g., through adjustments to formality, use of specific terms of address  (``dear''), or using in-group vocabulary. 
While prior computational work has studied the language of power and politeness for assessing hierarchies between speakers \citep[e.g.,][]{bramsen2011extracting,prabhakaran2012predicting,danescu2013computational,sap2017connotation}, little prior work on the language of intimacy exists, with most focusing on self-disclosure \cite{bak2012self,bak2014self}, which captures only a part of the concept of intimacy. 
In this study, we test the roles of two linguistic devices in intimate communication (\sref{sec:linguistic-devices}), hedging \cite{hyland2005metadiscourse} and swearing \cite{stapleton201012},  whose usages theory suggests should change relative to people's expressions of intimacy.

\myparagraph{Social Norms in Intimate Communications}
Group and sociocultural norms may strongly constrain the kinds of behaviors that are acceptable and desirable within certain situations \cite{allan1993social}. 
Frequently, these norms specify the acceptable levels of intimate interactions between people in specific social relationships and circumstances \cite{davies2013intergroup}, where violations of these norms lead to loss of face and social capital \cite{caltabiano1983variables}.
While types of relationships and closeness between people naturally build the expectations that certain levels of interactions are reserved for only selected social ties \cite{chelune1979self}, intimate behaviors and experience may not happen between people who are in close relationships \cite{hinde1981bases} and are thus regulated by larger social norms like gender. Societal views of gender roles significantly constrain the use of intimate communication, with specific expectations not only of the individual on the basis of their gender \cite{caltabiano1983variables} but dyadic effects depending on the gender identities present \cite{derlega1976norms}.\footnote{For example, prior studies have found that males who disclose very personal (intimate) information to other men are viewed as less well-adjusted \cite{derlega1976friendship} and are less well-liked \cite{lazowski1990self}.}
However, individuals are less adherent to these norms as they perceive themselves to be anonymous or when interacting with an individual whom they perceive they will not interact with again   \cite{rubin1983intimate,wynne1986quest,dindia1997self}; without the potential loss of face or social capital in such circumstances, individuals are more likely to engage in more intimate communication.
Thus, the norms of a social context and expectations around the loss of social capital for violations of these norms act as primary drivers of selecting the degree of intimacy expressed in a given context. Here, we test how intimacy varies across different types of social settings by varying dyadic gender composition in interactions (\sref{sec:gender}), social distance (\sref{sec:social-distance}), and perceived anonymity (\sref{sec:audience-design}).

\section{Quantifying Language Intimacy}
\label{sec:quantifying}
Questions provide a natural mechanism to study the intimacy of language.  In conversation, questions are frequently used to request information \cite{athanasiadou1991discourse}, providing the interlocutor with the opportunity to respond at a desired level of intimacy. This interactive questioning process can lead to the formation of intimate relationships as the subject matter and nature of disclosure increase over time \cite{aron1997experimental}. In this study, we aim to quantify intimacy in questions as a continuous variable because people naturally perceive intimacy along a continuum \cite{schaefer1981assessing}.  Following, we describe the dataset and annotation process for rating questions by intimacy.

\myparagraph{Data} 
Questions are drawn from 41 manually selected question-centered subreddits, e.g., \texttt{r/AskReddit}, which encompass a wide variety of topics and conversation styles.
The initial set of questions is derived from all post titles made in 2018 containing one question mark.
Then for each raw question, we remove Reddit-specific markup, e.g., ``[17M]'' or address terms to the community, e.g., ``Members of r/AskScience, $\ldots$'', replaced common abbreviations, e.g., ``AITA (Am I the Asshole)'' with their full expressions.
To ensure questions are self-contained, we require the question to be the entire post title and be a single sentence with at least four words. %
In total, this process yielded 3,212,969 questions; Appendices \ref{sec:subreddits} and \ref{sec:question_cleaning} contain the list of selected subreddits and question cleaning process.
From this dataset, we select 2,247 questions to annotate, balancing across months.

\myparagraph{Annotation} 
Rating the precise intimacy of question is a challenging task due to the potentially subjective nature of the question and the relative expectations of intimacy. Rather than directly estimating an intimacy value through scale-based annotation, we treat intimacy as a latent variable to be inferred from relative comparisons between questions. Following prior work in NLP on annotating social aspects of language \cite{bws-naacl2016}, we adopt a Best-Worst-Scaling (BWS) \cite{louviere2015best} scheme to estimate the latent intimacy values of questions. Here, four questions are shown as a tuple, and annotators are asked to identify the most intimate and least intimate questions of the tuple. As questions could be interpreted in multiple ways depending on the context, annotators were instructed to consider their judgments according to the expected intimacy if the question is asked in a scenario appropriate to its usage.
Each tuple annotation generates five pair-wise comparisons between questions' intimacy values that act as constraints when inferring the latent value on a continuous scale.

Prior to annotation, two annotators went through three rounds of training and discussed all disagreements. Following, annotators labeled an initial 212 tuples to assess exact agreement and subsequently, all other tuples were divided up between the two for annotation; these 8,563 tuples included 2,397 questions comprising 2,247 Reddit questions and an additional 150 questions from books, movies and Twitter for generalizability tests. Each question was presented in at least 12 tuples to ensure an accurate approximation.
To infer the latent intimacy values of all questions, we use Iterative Luce Spectral Ranking \cite{maystre2015fast} to convert the pair-wise comparisons into real-valued scores ranging from -1 (least intimate) to 1 (most intimate). 

To test the reliability of the ranked scores, we follow best practices \cite{kiritchenko2017best,mohammad2018obtaining} and compute the Split Half Ranking (SHR) by randomly splitting all the tuples into two sets, compute the intimacy scores within each, and compare the rankings; note that as the same questions appeared in both annotators' sets of tuples, the inferred ranks in each split reflect the judgments of both annotators. The Pearson's $r$ between the two sets' ranking scores is 0.776, which demonstrates high reliability in the annotations. See Appendices \ref{sec:annotation_guidline} and \ref{sec:data-sample} for annotation guidelines and data samples.

Additionally, annotators attained Krippendorff's $\alpha$=0.548 on 212 tuples. While this $\alpha$ is moderate as inter-annotator agreement (IAA) is normally measured, in BWS, lower agreement is expected when annotators encounter tuples where all four items are perceived to have essentially the same value, e.g., four factual questions asking nothing intimate; in such settings, annotators are likely to arbitrarily select the best and worst, which ultimately leads the items to have similar scores in the BWS scalar conversion process (as intended) but penalizes IAA. As a result, SHR is a better estimate of annotation quality and annotator reliability.

\section{Predicting the Intimacy of Questions}
\label{prediction}
Question intimacy is predicted using model-based regressors. We test two baseline models and two deep-learning regressors based on neural language models.  As baselines, we include two linear regression models with L2 regularization separately trained on either (1) bag of words features or (2) topic features. Bag of words features are constructed with unigrams, bigrams and trigrams. 
The second model uses an LDA model trained with 50 topics using Mallet\footnote{\url{http://mallet.cs.umass.edu/}} over a sample of 1M questions that includes the annotated questions; each question is then represented using its topic distribution for regression.
Our neural regressors use the RoBERTa \cite{liu2019roberta} language model as a base. We include two variants: one which is fine-tuned on 3M unannotated questions on a masked language modeling task, and a second which uses the default parameters in RoBERTa. Training uses only the 2,247 annotated Reddit questions, split 8:1:1 into training, validation, and test. %
Model settings and validation performance are listed in Appendix \ref{sec:model_details}.

\begin{table}[t!]

\newcommand{\tabincell}[2]{\begin{tabular}{@{}#1@{}}#2\end{tabular}}
\centering
\resizebox{0.49\textwidth}{!}{
\begin{tabular}{r cc}

\textbf{Model} &  \textbf{MSE} &  \textbf{Pearson's $r$} \\ 
\hline
Mean-value Predictor     & 0.08625 & 0.0000 \\
LR + Bag of Words        & 0.06532 & 0.5127 \\
LR + Topic Model & 0.05476 & 0.6211 \\
RoBERTa (base)           & 0.02855 & 0.8232 \\
RoBERTa (fine-tuned)     & \textbf{0.02106} & \textbf{0.8719} \\
\end{tabular}
}
\caption{   Question Intimacy Prediction Performance }
\label{model_comparison}
\end{table}

\myparagraph{Results}
Table \ref{model_comparison} shows that our best model, the fine-tuned RoBERTa model, attains a high correlation with human judgments as measured by Pearson's $r$. 
RoBERTa model with question fine-tuning outperforms the RoBERTa base model while both the RoBERTa models outperform all the other baselines. 
The topic model baseline is still able to attain moderate performance, matching the intuition that some topics are more intimate (e.g., romance) while others are less (e.g., mobile phones). However, as shown in Appendix \ref{sec:topic_analysis}, Figure \ref{fig:topci_plot}, many topics span the range of intimacies, demonstrating that estimating intimacy from topic alone is insufficient.

\myparagraph{Final Dataset}
To study question intimacy, we apply our fine-tuned RoBERTa regressor to our four different question datasets from Reddit, Twitter, books, and movies.
For Reddit data, we apply the same question extraction procedure to all content written in 2018 and extract all questions in posts and comments that receive a reply. %
This process yielded 16.6M post and 60.8M comment questions.

Twitter questions were collected from a 10\% sample of tweets from Jan 2018 to April 2020, where the tweet text was a single English question and was made as a direct message to a single person (i.e., reply or mention). We follow a similar question selection process as described in Section \ref{sec:quantifying}. %
Twitter questions were further processed by replacing all mentioned users (e.g., @StephenCurry30) with their screen names e.g., Stephen Curry), removing all  emojis, and removing all URLs.  %
After removing duplicates and self-replies, this process yielded 1.04M questions.

Book questions were collected from 51,224 English books on Project Gutenberg \cite{hart1992history}. BookNLP \cite{bamman2014bayesian} is used to identify characters' quotes and we identify 2.02M quotes ending with a question mark and having at least four words. %
We keep the full quote as a question, as the extended context was deemed necessary for correct interpretation.

Movie questions were extracted from the Cornell movie dialogue dataset \cite{danescu2011chameleons}, where all dialog lines  ending with a question mark and at least four words are treated as questions, which  yields 53,507 questions.

To test the generalizability of our model on these domains, the annotated data included 50 questions from each non-Reddit source, which were not included in the training data. %
Over this external dataset, our best-performing model achieved 0.6684, 0.6602, and 0.5233 Pearson's $r$ correlations on the intimacy ratings for book, Twitter, and movie questions, respectively. These moderately-high correlations demonstrate the generalizability of our model on outer domain data.

These four datasets allow detailed study on intimacy in language  and social factors due to their variety of content and social setting. Reddit and Twitter are social media data that contains real human messages, while book and movie data are imagined conversations that reflect social norms. Moreover, Twitter questions can be overlaid on its social network data to study the relationships between intimacy in language and social distance.

To test the reliability of our model prediction, the same annotators further annotated 300 question pairs sampled from the final dataset to reflect ranges of differences in their intimacy. Pairs of questions were grouped according to their difference in intimacy using a 0.1 range per group. 30 questions were sampled from each group.
Annotators selected which of the two questions was more intimate, or a third option if they had the same level of intimacy. 
Annotators attained Krippendorff's $\alpha$=0.70, indicating moderately-high agreement, with most disagreements happening for questions with small differences in intimacy as estimated by the model. Ultimately, 89\% of the question pairs have the same order for model prediction and human annotation, indicating the model's estimates of intimacy do match human judgments.

\section{Intimacy and Pragmatic Choices }
\label{sec:linguistic-devices}

In language, individuals can construct intimacy through stylistic choices that signal their view of the world and personal relationship to the proposed ideas \cite{bell1984language}. When questions carry the risk of losing face---e.g., broaching more intimate topics beyond what is socially acceptable in the current context---individuals reduce their commitment to the act through linguistic mitigation \cite{fraser1980conversational}. Here, we connect interpersonal pragmatics to the language of intimacy, showing how individuals perform pragmatic acts in their questions to mitigates risk as intimacy increases, much like how politeness is employed to save face \cite{brown1987politeness}.
In particular, we examine pragmatic choices in questions around 
(i) the speaker's certainty, expressed in hedges from \citet{hyland2005metadiscourse} (ii) the speaker's belief of the social distance, expressed in swearing.
To analyze these choices, we compare the mean intimacy ratings in questions relative to whether a specific  strategy is employed; to ease comparison across datasets, we first standardize intimacy ratings within each domain.

\begin{figure}[t]
\includegraphics[width=3.1in]{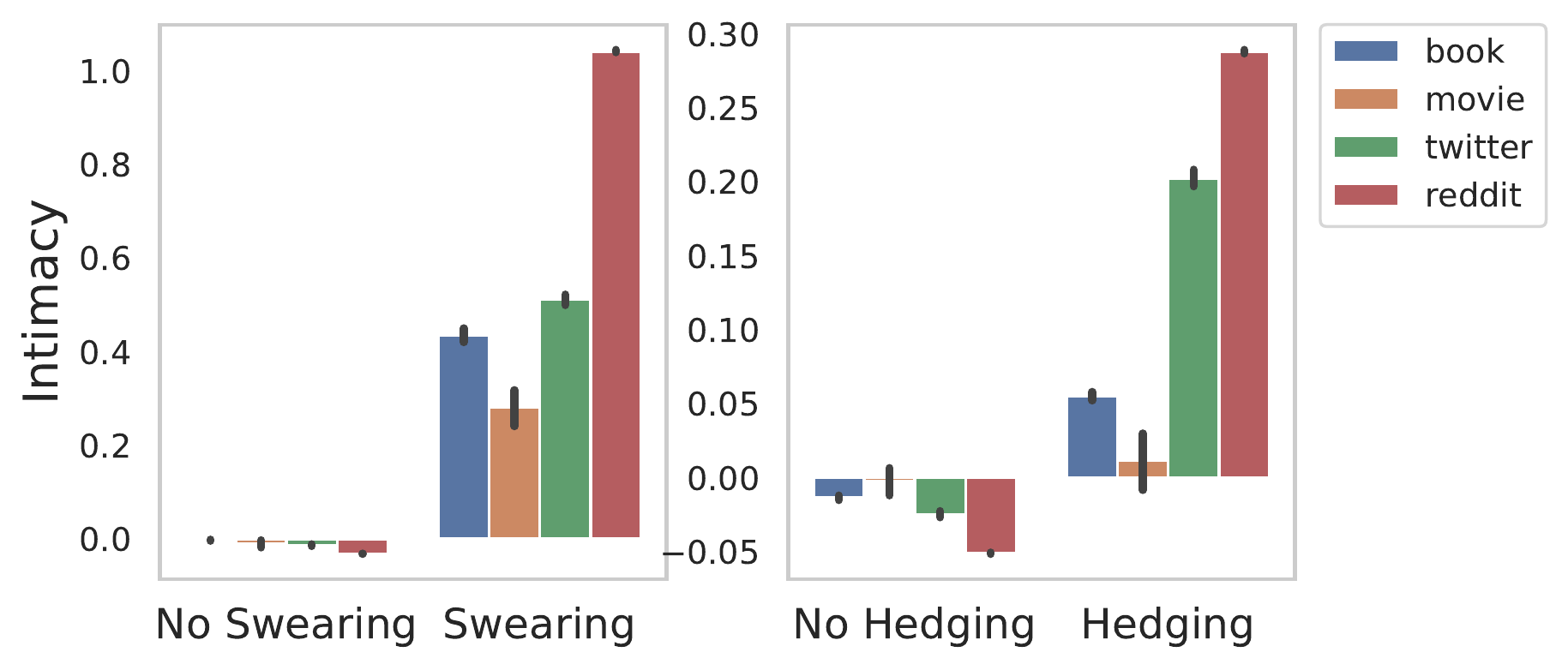}
  \caption{Relative levels of intimacy for questions containing socoiopragmatic markers show that individuals use swearing in intimate settings likely as a way of indicating closeness and hedge when asking intimate questions to decrease risk. Intimacy values are standardized within domain for comparability.}
  \label{linguistic_cues}
\end{figure}

\myparagraph{Certainty}
Hedging is a marker of intentional vagueness \cite{lakoff1975hedges} and aims to reduce risks in interpersonal communications \cite{caffi1999mitigation}.  Expressing uncertainty in a question can allow people to ask more intimate questions  without the risk of inappropriateness. For example, in the following two questions, the hedge (i.e., might) serves to allow the respondent to answer with uncertainty or vagueness, reducing the risk from forcing an overly-intimate answer. 
\begin{enumerate}[label=(\alph*)]
\item What \textbf{might be} your best childhood memory?
\item What is your best childhood memory?
\end{enumerate}
Figure \ref{linguistic_cues} (Right) shows that questions containing hedging words are generally more intimate than other questions, which is highly consistent across different domains. This result indicates people regularly employ hedging as a strategy  to reduce risk when asking more intimate questions.

\myparagraph{Social Distance}
While the use of swear words explicitly transgresses social norms  \cite{andersson1990bad,monaghan2012cultural}, the act of swearing can express the speaker's perceived solidarity with the audience \citep[][p.~296]{stapleton201012}. In this discursive act, the intentional act of swearing emphasizes in-group status with the audience and normalizes the use of words that would be taboo to out-group members \citep[][p.~99]{fagersten2012s}. Therefore, when asking questions, swearing may be employed to construct the perception of stronger social ties that would license more intimate questions.
Large-scale analysis across domains also supports this hypothesis. Figure \ref{linguistic_cues} (left) shows that questions containing swear words are far more intimate than others. This finding is consistent in both real conversations (Twitter, Reddit) and imagined conversations (movies, books).

\section{Gender Norms in Language  Intimacy }
\label{sec:gender}

Gender is one of the earliest learned social norms for individuals \cite{west1987doing,martin2010patterns}, with strong gender expectations around intimacy in conversation \cite{caltabiano1983variables}.
Social psychologists have found that women show more interest in verbal intimacy than men \cite{blumstein1983american,engel1986love}, and are more likely to initiate intimate verbal interactions in marriages \cite{markman1989men}. Even in friendship, female friendships typically involve  more intimate self-disclosures than male friendships \cite{aries1983close,davidson1982friendship,lewis1978emotional}.

Our four datasets provide an ideal setting for testing theories of gender expectations of intimacy along two fronts. First, relatively-anonymous social media like Reddit provide few social cues about the identity of the person; in these deindividuated settings, do gender norms persist?
Second, film and literature reflect imagined conversations that require authors to ``do gender'' \cite{west1987doing} from their internalized expectations around intimacy, which is not regulated by actual loss of face for norm violations. In these imagined settings, do authors perform  gender expectations on their characters, and are expectations consistent for authors of a different gender?

\myparagraph{Methods} 
A user's gender\footnote{Gender is a complex social construct beyond male and female and we acknowledge the known presence of a small number of nonbinary and transgender individuals in our fictional works, as well as their likely presence in Twitter and Reddit. However, we were regrettably unable to reliably identify these gender identities using current techniques.} in social media is inferred from their username using GenderPerformer \cite{wang2018s}, which was trained to operate on social media like Reddit and only returns a gender label for usernames that strongly perform male or female. In movie scripts, we use both the gender labels provided for 3,015 characters in the Cornell movie dialogue dataset \cite{danescu2011chameleons} and a second approach to infer gender for another 2,872 characters using a name database based on US baby names from 1930-2015.
For questions in books, BookNLP \cite{bamman2014bayesian} is used to identify the speaker of each question using coreference resolution to identify a canonical name; the speaker's name is then matched using US census names and checked against gendered titles (e.g., Mr.) or roles (``mother''). Additional details on the matching process are provided in Appendix \ref{sec:gender_inference}.

To test for differences in intimacy norms for authors, we construct separate mixed-effect regressions to predict the intimacy of the question for male and female authors. 
Each domain may have its own level of intimacy, therefore we standardize all intimacy scores within domain to compare z-scores across domain when examining the effect of dyadic gender composition.
Each regression includes a fixed effect for the gender of the speaker and audience and nested random effects for the author and book. These random effects effectively control for idiosyncratic differences in authors' perceptions of intimacy, relative differences across genres, and the time period in which the book was written. From this regression, we estimate the average marginal effect on intimacy for depicting a particular gender composition of the dyad, using female-female as the reference category.

\myparagraph{Results} 
\begin{figure}[t]
\centering
\includegraphics[width=3.0in]{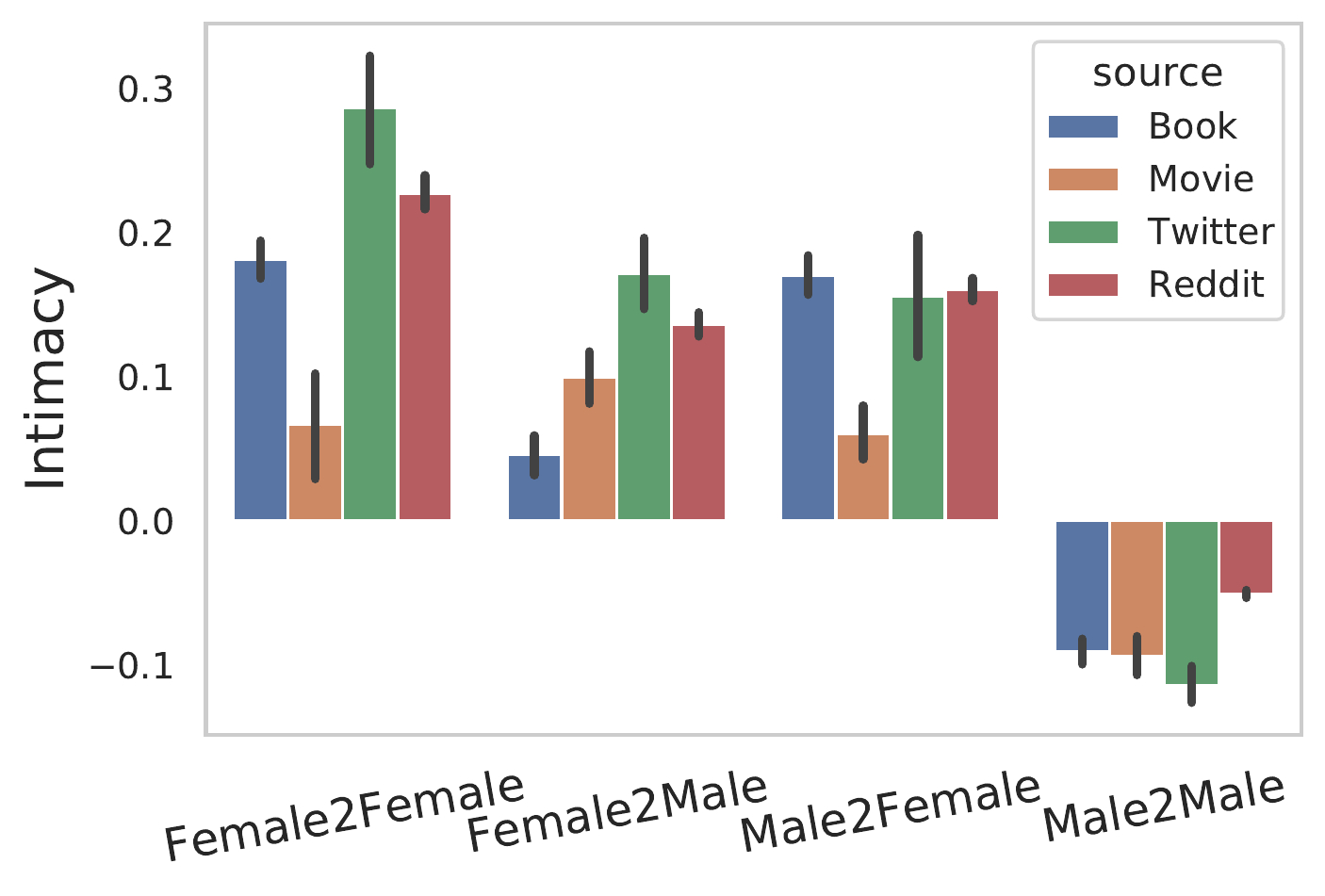}
  \caption{Relative levels of intimacy for questions show that male-to-male conversations hold the least intimacy among four gender dyads. This pattern persists across both real social media data and imagined conversations in books and movies, which indicates a strong social norm of masculinity. 
  }
  \label{fig:gender_diff}
\end{figure}
Dyadic interactions in all four datasets---real and imagined---follow expected social norms for gender and intimacy (Figure \ref{fig:gender_diff}). Although the relative intimacy levels differ across datasets, female-female questions were the most intimate and the presence of a female audience licenses males to ask more intimate questions, on par with those of females. In contrast, male-male dyadic interactions follow the low-intimacy hegemonic norms of masculinity, where men are supposed to be strong, rational, and inexpressive of personal emotions \cite{edwards2004cultures,donaldson1993hegemonic}.

\begin{figure}[t]
\centering
\includegraphics[width=3.0in]{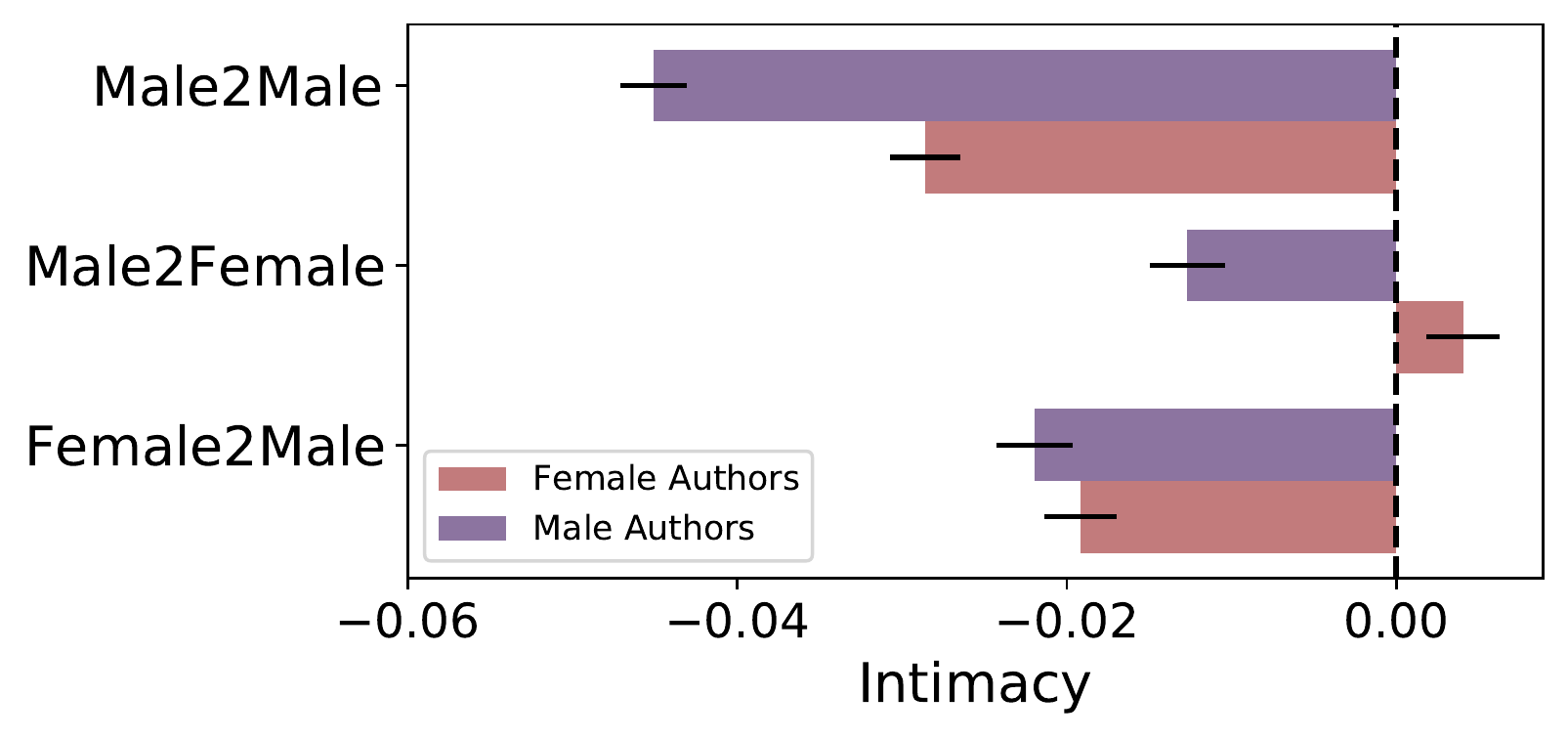}
  \caption{Averaged marginal effect of different gender dyads on  intimacy grouped by author gender. Female-to-Female is the dashed baseline here. Both male and female authors write less intimate male-to-male language, which  mirrors expectations of lower intimacy.}
  \label{fig:a_gender}
\end{figure}

Do female authors also perpetuate gendered intimacy norms of males or are the trends in \fref{fig:gender_diff} driven by male authors only? Shown in Figure \ref{fig:a_gender}, these norms persist regardless of whether the interaction is described by a male or female author: male-male interactions have the lowest intimacy in conversation. This result suggests that despite female authors not having direct experience with such interactions, normed expectations around gender intimacy are so firmly established that  they persist across gender in imagined settings---even when controlling for genre and time period. 
However, the disparity between male-male interactions and others is highest for male authors, suggesting these authors reinforce this norm more strongly.
Full regression details are in Appendix \ref{regression}.

\section{Social Distance and Intimacy}
\label{sec:social-distance}

The appropriateness of a specific level of intimacy and associated cost for transgressing expectations vary depending on the social expectations. Among close friends, intimate questions are a natural form of discourse and carry low social risk \cite{dosser1983situational,miller1990intimacy}. 
However, people may also share very intimate information with strangers \cite{simmel1950sociology,rubin1975disclosing}, commonly referred to as the \emph{strangers on a train} effect \cite{rubin1983intimate}. Individuals in these encounters have little likelihood of future interactions, removing the consequences for violating intimacy norms around increased disclosure \cite{thibaut2017social,wynne1986quest}.
In contrast to both friends and strangers, individuals are least intimate with casual acquaintances for which there are some expectations of potential future interaction and, therefore, longer-term consequences for norm violations. Together these behaviors point to a hypothesized U-shaped relationship between intimacy and social distance in in-person settings \cite{rubin1975disclosing}.
In social media, individuals come in contact with all three of these cohorts and have the potential to regularly connect individuals with complete strangers. Here, we ask whether these offline behaviors translate to a ``strangers on the internet'' phenomenon.

\begin{figure}[t]
\centering
\includegraphics[width=2.9in]{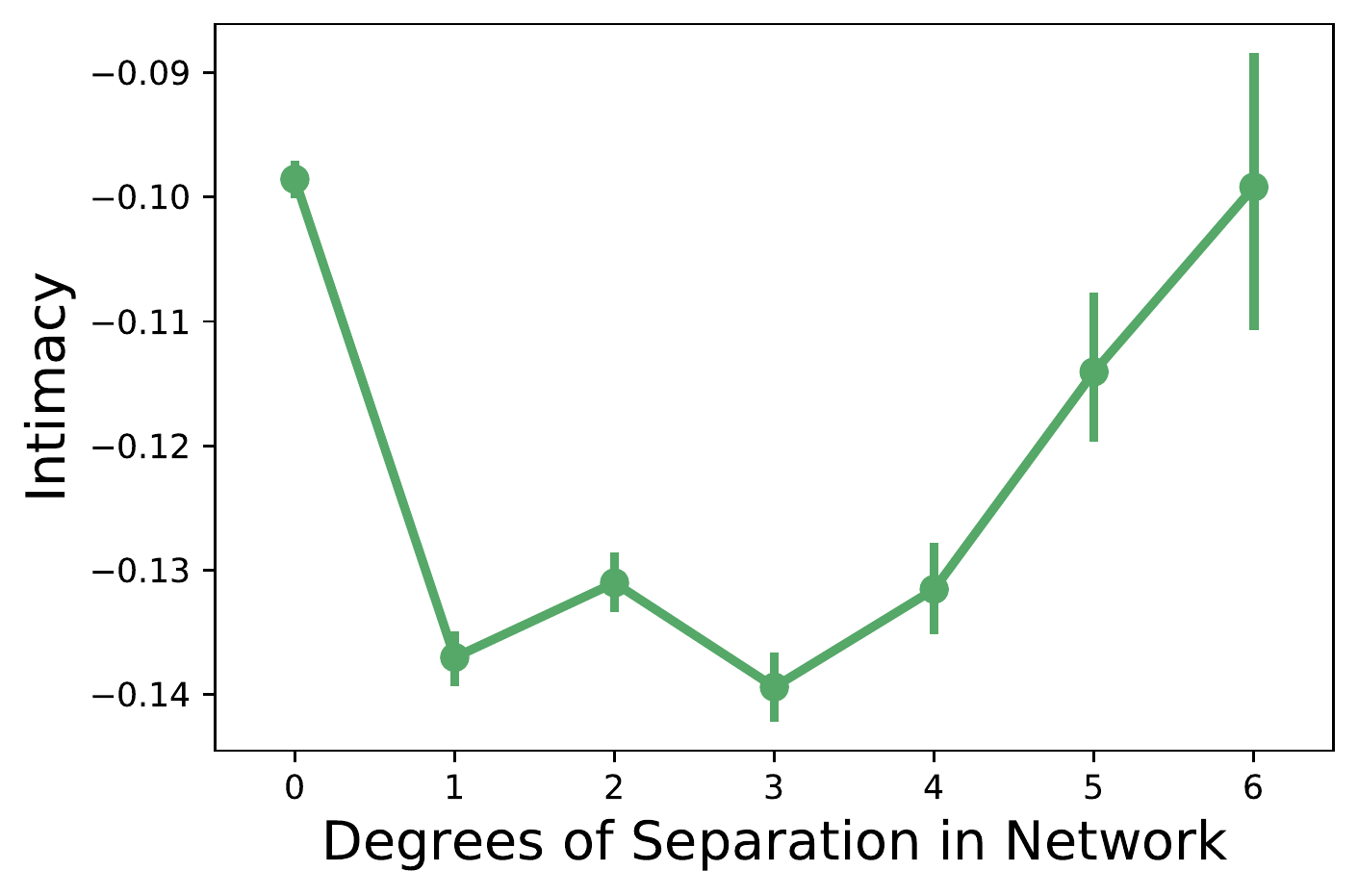}
  \caption{Intimacy in language and the social distance between users. The most intimate interactions happen between close friends or total strangers.}
  \label{fig:shortest_path}
\end{figure}

\myparagraph{Data and Methods}
To estimate familiarity and expected future contact between individuals, we construct a social graph from Twitter of all mentions from a 10\% sample of tweets made from January 2013 to April 2020. Following common practice \citep[e.g.,][]{jurgens2013thats,myers2014information,faralli2015large}, individuals are considered to have a relationship if they both mention each other. The resulting graph contains 1.1B edges. 

For each question tweet in our dataset, we measure the degrees of separation (path length in the graph) between the question-asker and recipient. Recipients with an immediate social relationship have a degree of 0. As the network is constructed from a 10\% sample, our estimates of degree contain Type II bias and may overestimate the degree (e.g., by not seeing an interaction); however, many individuals do ask questions to complete strangers through encounters on shared discussions (e.g., around a hashtag). To minimize confounds due to user popularity (e.g., celebrities and politicians), we remove all tweets directed  to verified accounts or those with $\ge$5000 followers.

\myparagraph{Results} 
As shown in Figure \ref{fig:shortest_path}, intimacy behavior on Twitter mirrors the U-shaped curve predicted from offline experiences \cite{simmel1950sociology,rubin1975disclosing}, where people ask the most intimate questions to close ties and complete strangers, with a trough for acquaintances where norm violations have the highest cost.  
Further, individuals ask strangers questions with the same level of intimacy as friends, but  these strangers must be very distant in the network; our results suggest that individuals are highly sensitive to the perceived risk of future interaction with lower intimacy rates even four degrees of separation away.
In in-person settings, psychologists have largely been unable to measure the exact degree of separation between people due to the cost and difficulty of such a large-scale experiment; using a global social network, our result provides the first quantitative estimate of the relationship between distance and intimacy. 

\section{Anonymity as Audience Design}
\label{sec:audience-design}

Social media creates a new affordance for side-stepping the norms around intimacy: anonymity. By communicating through an anonymous account, an individual ensures that they are viewed as a stranger, removing the social cost of norm transgressions around intimacy for gender and social distance.
Prior work has shown that the use of anonymous accounts is not necessarily driven by their willingness to publicly disclose, but rather around perceived anonymity and privacy as a way of performing identity and boundary management \cite{de2014mental,leavitt2015throwaway}. 

Individuals shift their language based on the expected audience, with  \citet[][p.~185]{bell1984language} noting that these shifts can ``simulate or create intimacy with a stranger;'' social media complicates this audience design process through its context collapse \cite{marwick2011tweet} where individuals must choose content and style to simultaneously match the norms and expectations of their different social circles \cite{androutsopoulos2014languaging}.
Given an audience of an unknown composition, individuals may be inhibited from style-shifting into more intimate language due to the perceived risk of social capital loss.
However, full anonymity could free the speaker from the penalty of norm violations, allowing them to shift to a desired intimacy level without risk. %
Following, we test to what degree does anonymity facilitate increased intimacy. 

\myparagraph{Methods}
Anonymous accounts were collected by identifying posts made in 2018 on Reddit by usernames containing \emph{throwaway} or \emph{anonymous}, which are recognized markers of intentional anonymity on Reddit \cite{leavitt2015throwaway}. 
The intimacy of language by \textbf{Anonymous} accounts is compared relative to three groups: (1) accounts containing a  first name in the username, e.g., \texttt{SamIsCool}, as these potentially signal a closer association with personal identity, which we refer to as \textbf{Name Containing}; (2) accounts without any explicit demographic or identity marker, e.g., \texttt{atomiccyle}, which are referred to as \textbf{Depersonalized}, and (3) all other accounts within a subreddit. Details on name and demographic matching are provided in Appendix \ref{identity-lexicon}.
A total of 12,528,813 questions were collected across 117,526 subreddits.
We fit a mixed-effect regression to predict the intimacy of the question from the identity presentation of the author, using random effects for each subreddit to control for different levels of intimacy in each; the all-other category of names is treated as the reference group in categorical coding.

\begin{figure}[t!]
\centering
\includegraphics[width=3.0in]{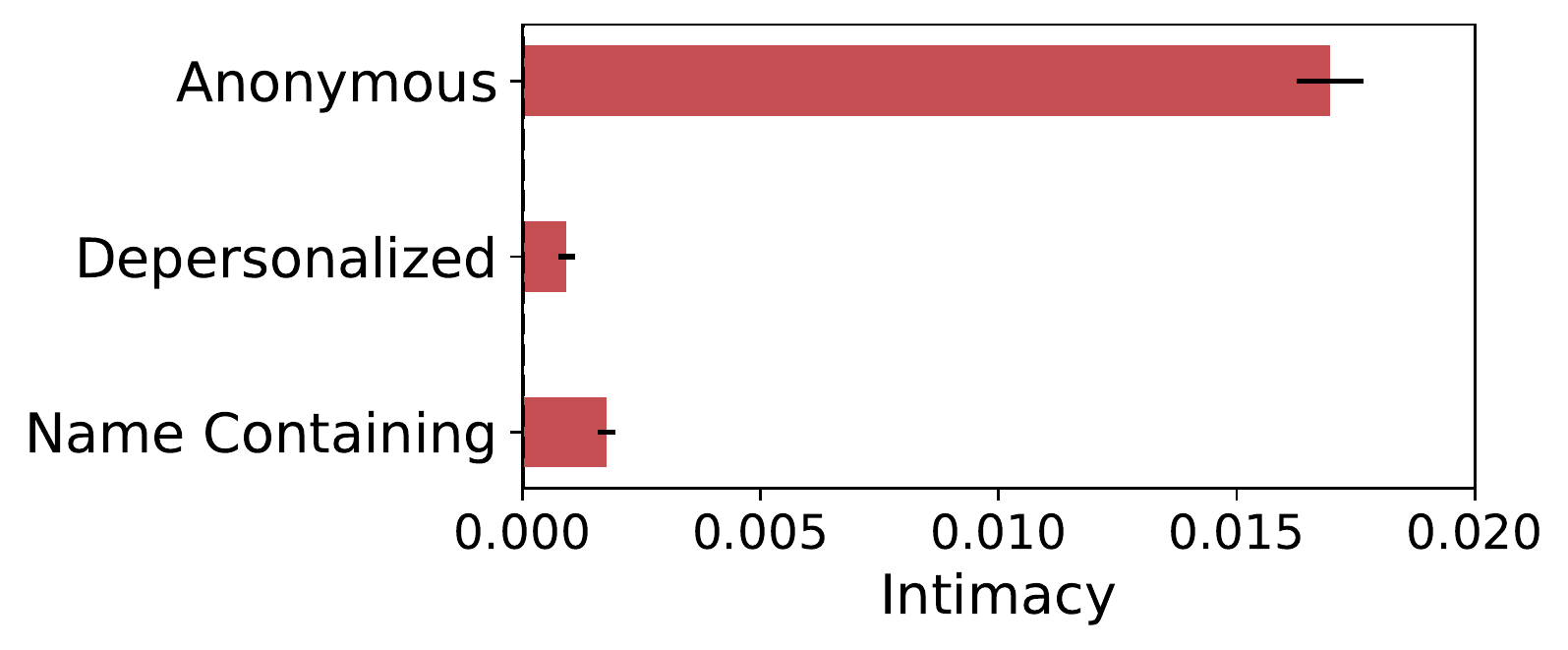}
  \caption{Averaged marginal effect on intimacy in language for specific types of Reddit accounts, relative to all other accounts as the reference category, shows that anonymous accounts have higher intimacy in their language than other accounts in the same communities.}
  \label{fig:anonymity}
\end{figure}

\myparagraph{Results}
Anonymous users ask substantially more intimate questions than any other types of accounts (Figure \ref{fig:anonymity}), even when controlling for the social context of those questions (via subreddit random effect). 
However, all other users ask substantially less-intimate questions, even if their username is effectively anonymous; model coefficients are listed in Appendix \ref{regression}. This result points to the perceived loss of face even for otherwise-anonymous users who may have a reputation on the platform. Only through explicit anonymity (e.g., a throwaway account) do users substantially violate the contextual social norms around intimacy in a community.
From a linguistic perspective, creating a separate anonymous identity to ask intimate questions can be viewed as a special strategy of audience design. Rather than changing the style of the expression to match an audience, anonymity enables changing the social cost of the desired style.

\section{Discussion and Ethics}

This work has focused on analyzing intimacy in language through questions, showing consistent findings across four studies of how individuals modulate intimacy in their communication with respect to the norms of their social surroundings. Although questions are only one part of language, they represent a natural starting point due to their interpersonal nature and our findings open the door to broader studies on other types of communication.
Further, our work has applications in many NLP settings. For example, intimacy measurements can provide a useful metric for context-sensitive offensive language detection; given an ongoing conversation, a question that is substantially more intimate than normal suggests that it might be offensive (or at least violate social norms). Our models would allow tracking intimacy changes to separate offensive questions from those in conversations that gradually become more intimate.
As a second example, dialog systems can benefit from intimacy models through adjusting their language to match user preferences---or potentially encourage interactions that lead to more intimate topics.

The study of intimacy in language necessitates a discussion of ethical choices and implications. All experiments were performed on public data, in accordance with terms of service; as users of social media have contextual expectations of privacy, all examples of questions and usernames in this paper have been paraphrased to preserve anonymity. One risk posed by our technology is using these models to seek out especially-intimate questions from users in order to abuse or embarrass them.  As one potential mitigation, platforms might use this same technology to prompt users to switch to a throw-away account when asking the question.

\section{Conclusion}

This paper represents a step towards a full understanding of the social information in language through new data and models for studying intimacy in language. By developing a high-quality dataset of questions rated for their intimacy and a corresponding model that closely correlates with human judgments, we study 80.5M questions across social media, books, and movies to reveal how individuals shape and react to their social setting through selecting the intimacy of their language.
In four studies, we show that the intimacy of language is not only a personal choice, where people may use different linguistic strategies for the expressions of intimacy but reflects constraints from social norms, including gender and social distance. Our study provides strong evidence for existing findings in social psychology and also enriches the study of computational sociolinguistics in NLP community.

\section*{Acknowledgments}\label{sec:ack}
We thank anonymous reviewers and area chairs for their helpful suggestions.  This material is based upon work supported by the National Science Foundation under Grants No 1850221 and 2007251. The last author was supported in part through the Amazon Alexa Prize Competition 3.

\bibliography{references}
\bibliographystyle{acl_natbib}

\appendix

\section{List of Question Centered Subreddits}
\label{sec:subreddits}
For data annotation and language model fine-tuning, we use questions sampled from the following subreddits:

OutOfTheLoop
IWantToLearn
whatisthisthing
answers
NoStupidQuestions
amiugly
whatsthisbug
SampleSize
TooAfraidToAsk
whatsthisplant
IsItBullshit
morbidquestions
ask
AskReddit
shittyaskscience
TrueAskReddit
AskScienceFiction
AskWomen
AskMen
askgaybros
AskRedditAfterDark
asktransgender
AskMenOver30
askscience
AskHistorians
AskCulinary
AskSocialScience
AskEngineers
askphilosophy
AskDocs
explainlikeimfive
ExplainLikeImCalvin
relationships
relationship\_advice
legaladvice
bestoflegaladvice
Advice
AmItheAsshole
MechanicAdvice
needadvice
dating\_advice

\section{Gender Inference}
\label{sec:gender_inference}
User gender in social media (i.e. Twitter and Reddit) is inferred from the username using GenderPerformer \cite{wang2018s}, which was trained to operate on social media like Reddit. In movie scripts, we use both the gender labels provided for 3,015 characters in the Cornell movie dialogue dataset \cite{danescu2011chameleons} and a second approach to infer gender for another 2872 characters using name database based on US baby names from 1930-2015. \footnote{\url{https://www.ssa.gov/oact/babynames/limits.html}} Such a strategy has been widely used in previous works \cite{west2013role,prabhakaran2014gender}. 
For questions in books, BookNLP \cite{bamman2014bayesian} is used to identify the speaker of each question, and we follow the similar name matching strategy for movie questions to recognize the gender of speakers. For recipients, we look for addressee information using regular expressions. For example, for the question "What is this, Tom?", we first extract \texttt{Tom} using regular expressions to match words between ``,'' and ``?'', and then use the gender name database to identify the gender. If the word is not found in the database, we secondly check gender special words (e.g., he, wife, sister) for book questions.\footnote{Male: man, he, Mr., boy, husband, him, uncle, guy, sir, brother, father; Female: woman, she, Mrs., Miss, girl, madam, her, aunt, wife, sister, mother}

Please note that we believe non-binary genders and transgenders are also vitally important and valuable for intimacy research. However, for this current work, we 
only identify binary genders following common practices and leave the study of other non-binary genders and transgenders in future research.

\section{Identity Lexicon}
\label{sec:identity_lexicon}
\paragraph{Anonymous} accounts contain strings indicating anonymous identity including: \texttt{anonymous}, \texttt{anon} and \texttt{throwaway}. For \texttt{anon} we also require the username to end with digits. We use regular expressions to mach all the usernames meeting the criteria above.

\paragraph{Name Containing} accounts contain real-world names,\footnote{US baby names in 2016 \url{https://www.ssa.gov/oact/babynames/limits.html}} which are treated as a marker of identity. Here, we restrict names to be CamelCased or containing special symbols (i.e. \texttt{-} and \texttt{\_}). Moreover, some names in the database might be primarily used as other functions instead of names (e.g. \texttt{rainbow} and \texttt{my}), to eliminate the potential bias, we manually checked 500 most frequently matched names and removed those might be used in context other than names.

\paragraph{Depersonalized} accounts are those without common demographic markers including: gender, age, socioeconomic info, religion and political identity. We select accounts marked as ungendered by GenderPerformer.\footnote{\url{https://pypi.org/project/genderperformr/}} Then we further remove account names whose suffix likely denotes some form of age information by identifying usernames ending with 4 digits from 1950 to 2005 and 2 digits from 50 to 99. After this, we also removed accounts containing lexicons from three other categories using regular expressions. For lexicons containing less than 4 letters, we ensure that only when they are a subsplit of CamelCased string or usernames connected by \_ will they be identified. Here is the list of identity lexicons.

\begin{enumerate}
    \item Political: briebart, rightwing, imstillwithher, im\_still\_with\_her, left\_wing, lockherup, obama, bernie\_sanders, maga, leftie, leftwing, neocon, liberal, republicans, republican, libtard, democrap, democrats, trump, conservative, im\_with\_her, imwithher, bernie, neo conright\_wing, democratic, lock\_her\_up, democrat, clinton
    \item Religions: allah, lutheran, atheist, bible, buddah, jewish, christ, muslim, islamic, buddhism, jesus, shariah, catholic, buddhist, quran, torah, buddha, methododist, christianity, athiest, athiesm, judaism, koran, jew
    \item Socioeconomic: mdphd, phd, dumb\_hick, ghetto\_fabulous, hillbilly, boondocks, hill\_billy, yokel, yokels,  lawyer, ghetto, hillbillies, hayseed, hayseeds, rednecks, professor, backwoods, beer\_drinkin, ghettofabulous, bumpkins, prof, dphil, red\_neck, redneck, beerdrinkin, beerswillin, bumpkin, doctor, dds, bubbas
\end{enumerate}
\label{identity-lexicon}

While these lexicons are by no means exclusive to the types of identities a person might signal in their username, they still provide some utility for contrasting the behaviors of users who do chose to identify these sociodemographic signals with those that do not (e.g., pizzamagic).

\section{Hedge Words and Swear Words}
\label{sec:hedge_swear_words}
The linguistic analysis of intimacy in Section 5 of the main paper uses two existing lexicons.
For hedge words, we use the list provided by \newcite{hyland2005metadiscourse}, which comprises 100 common hedge words in scientific writing. For swear words, we use the swear word list\footnote{\url{https://github.com/RobertJGabriel/Google-profanity-words/blob/master/list.txt}} used by Google, which covers a wide range of swear words.

\begin{table}[t!]

\newcommand{\tabincell}[2]{\begin{tabular}{@{}#1@{}}#2\end{tabular}}
\centering
\resizebox{0.49\textwidth}{!}{
\begin{tabular}{r ccc}

\textbf{Model} &  \textbf{MSE} &  \textbf{Pearson $r$} & \textbf{Training time} \\ 
\hline
Mean-value Predictor     & 0.08625 & 0.0000 &  $<$ 1s \\
LR + Bag of Words        & 0.06532 & 0.5127 &  $<$ 1s \\
LR + Topic Model & 0.05476 & 0.6211 & $<$ 1s \\
RoBERTa (base)           & 0.02855 & 0.8232 & $<$ 10s/epoch\\
RoBERTa (fine-tuned)     & \textbf{0.02106} & \textbf{0.8719} & $<$ 10s/epoch\\
\end{tabular}
}
\caption{Performance on the validation set at estimating intimacy and the training times for each model. }
\label{Vlidation details}
\end{table}
\section{Model Details}
\label{sec:model_details}
We use scikit-learn version 0.23.1 to build the linear regression model \cite{pedregosa2011scikit}. Specifically, for the linear model, we use ridge regressor with default settings.  The built-in CountVectorizer of scikit-learn is used to vectorize the unigram, bigram and trigram of each input question. The size of the bag-of-words feature vector is set as 10000.

For all the RoBERTa models \cite{liu2019roberta}, we use Hugging Face\footnote{\url{https://huggingface.co/}} transformers and set the batch size as 128 and learning rate as 0.0001. We set $max\_len=50$. Adam \cite{kingma2014adam} is used for optimization. All the other hyperparameters and the model size are the same as the default \texttt{roberta-base} model.\footnote{\url{https://github.com/pytorch/fairseq/tree/master/examples/roberta}} We train both the model for 30 epoches and choose the model with lowest MSE on validation set. For the question fine-tuning process, we simply follow all the default settings recommended by Hugging Face. Regarding hyperparameter trials, we only tuned the learning rate as 0.001, 0.0001 and 0.00001. We found that 0.001 didn't lead to a good performance while 0.0001 and 0.00001 both achieved good scores regarding MSE and Pearson r. So we simply go with 0.0001 for both the RoBERTa models. All the code, datasets and parameters of our best-performing model are released and one could easily reproduce all the experiments.

\section{Additional Regression Results}
\label{regression}

Here, we show the regression results of two analyses for the intimacy in different gender compositions of a dyad (Table \ref{authro_gender_r}) and how the relative anonymity of one's account name predicts the intimacy of the question that is asked (Table \ref{anonymity_r}). These tables show the model coefficients and standard errors for the mixed-effect regressions described in the main paper; the figures in the paper reflect the bootstrapped average marginal effects of the relevant categorical variable.

\begin{table}[!tbp] \centering 
 \resizebox{0.49\textwidth}{!}{
\begin{tabular}{@{\extracolsep{5pt}}lcc} 
\\[-1.8ex]\hline 
\hline \\[-1.8ex] 
 & \multicolumn{2}{c}{\textit{Dependent variable:}} \\ 
\cline{2-3} 
\\[-1.8ex] & \multicolumn{2}{c}{intimacy} \\ 
 & Female Author & Male Author \\ 
\\[-1.8ex] & (1) & (2)\\ 
\hline \\[-1.8ex] 
 Female-to-Male & $-$0.019$^{***}$ & $-$0.022$^{***}$ \\ 
  & (0.002) & (0.002) \\ 
  & & \\ 
 Male-to-Female & 0.004$^{*}$ & $-$0.013$^{***}$ \\ 
  & (0.002) & (0.002) \\ 
  & & \\ 
 Male-to-Male & $-$0.029$^{***}$ & $-$0.045$^{***}$ \\ 
  & (0.002) & (0.002) \\ 
  & & \\ 
 \textit{intercept}& 0.050$^{***}$ & 0.057$^{***}$ \\ 
  & (0.002) & (0.002) \\ 
  & & \\ 
\hline \\[-1.8ex] 
Observations & 41,569 & 66,862 \\ 
Log Likelihood & 17,951.830 & 28,731.180 \\ 
Akaike Inf. Crit. & $-$35,889.670 & $-$57,448.370 \\ 
Bayesian Inf. Crit. & $-$35,829.220 & $-$57,384.590 \\ 
\hline 
\hline \\[-1.8ex] 
\textit{Note:}  & \multicolumn{2}{r}{$^{*}$p$<$0.1; $^{**}$p$<$0.05; $^{***}$p$<$0.01} \\ 
\end{tabular} 
}
\caption{\label{authro_gender_r} Fixed effect regression analysis of different gender dyads on linguistic intimacy grouped by author gender, where an individual with one gender speaks to an individual with another gender. Nested effect of author and books are controlled and Female-to-Female is the reference category. All the results except for Male-to-Female dyad of Female author are statistically significant.}
\end{table}

\begin{table}[t!] \centering 
 \resizebox{0.49\textwidth}{!}{
\begin{tabular}{@{\extracolsep{5pt}}lc} 
\\[-1.8ex]\hline 
\hline \\[-1.8ex] 
 & \multicolumn{1}{c}{\textit{Dependent variable:}} \\ 
\cline{2-2} 
\\[-1.8ex] & intimacy \\ 
\hline \\[-1.8ex] 
 Anonymous & 0.017$^{***}$ \\ 
  & (0.001) \\ 
  & \\ 
 Depersonalized & 0.001$^{***}$ \\ 
  & (0.0002) \\ 
  & \\ 
 Name Matched & 0.002$^{***}$ \\ 
  & (0.0002) \\ 
  & \\ 
 \textit{intercept} & $-$0.213$^{***}$ \\ 
  & (0.0005) \\ 
  & \\ 
\hline \\[-1.8ex] 
Observations & 12,528,813 \\ 
Log Likelihood & 3,076,780.000 \\ 
Akaike Inf. Crit. & $-$6,153,549.000 \\ 
Bayesian Inf. Crit. & $-$6,153,463.000 \\ 
\hline 
\hline \\[-1.8ex] 
\textit{Note:}  & \multicolumn{1}{r}{$^{*}$p$<$0.1; $^{**}$p$<$0.05; $^{***}$p$<$0.01} \\ 
\end{tabular} 
}
\caption{\label{anonymity_r} Fixed effect regression analysis of different types of Reddit accounts regarding question intimacy, controlled for subreddit; usernames categorized as Other are used as the reference category. Anonymous accounts are asking more intimate questions than other accounts, indicating an anonymous identity is in part of the users' audience design to initiate more intimate interactions. All the results are statistically significant.}
\end{table}

\section{Question Cleaning Rules}
\label{sec:question_cleaning}
Reddit questions can potentially contain significant noise from Markdown, Reddit-specific jargon, or the otherwise-noisy nature of social media. To avoid training our model on such data, we adopted the following pipeline, shown in Table \ref{question_cleaning}, to either exclude or modify questions prior to inclusion. Table \ref{question_cleaning} also includes examples of the resulting modifications.

\begin{table*}[h!]
\small

\newcommand{\tabincell}[2]{\begin{tabular}{@{}#1@{}}#2\end{tabular}}
\begin{center}
 \resizebox{0.99\textwidth}{!}{
\begin{tabular}{l|l|l}
\toprule
\textbf{Rule} & \textbf{Example} & \textbf{Result}\\
\midrule
Remove multiple sentences or sentence without ending marker & SO many chaos & \textit{Removed} \\
Remove sentences without question mark & You are not saying this & \textit{Removed} \\
Replace multiple markers with one question marker & Why are you doing this !!!!? & Why are you doing this ? \\
Remove inserted meta-information in questions & My husband[30M] ... & My husband ... \\
Replace special abbreviations with its full expressions & AITA in doing this? & Am I the Asshole in doing this ?\\
Replace html symbols  & $\&amp\;$ & and\\
Remove questions with fewer than four words & That thing? & \textit{Removed}\\

\bottomrule

\end{tabular}
}
\caption{ The sequential set of rules applied top to bottom for determining if a question asked in Reddit is included in our dataset.}
\label{question_cleaning}
\end{center}
\end{table*}

\section{Annotation Guidelines and Preparation}
\label{sec:annotation_guidline}
\begin{figure*}[t]
\centering
\includegraphics[width=6.5in]{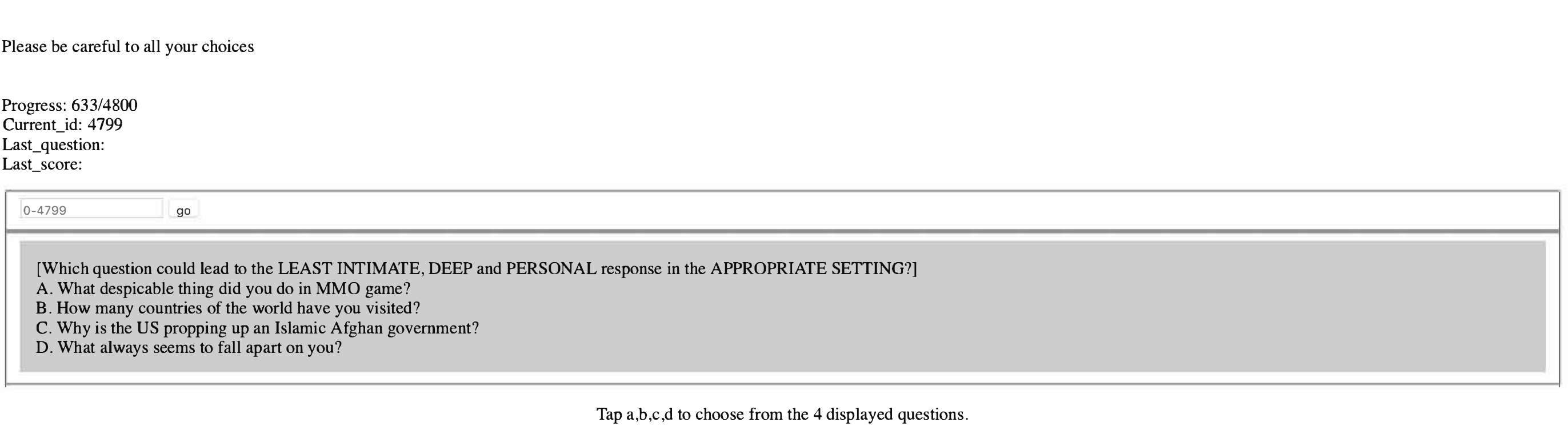}
  \caption{User Interface for Intimacy Annotation}
  \label{fig:annotation}
\end{figure*}

Each annotator is asked to choose the question that could ``lead to the Most/Least INTIMATE, DEEP and PERSONAL response in the APPROPRIATE SETTING'' among four randomly selected questions. Figure \ref{fig:annotation} shows the user interface of our web-based annotation tool. The authors conducted several rounds of pilot annotation trials among seven annotators, prior to beginning annotation for the current study's data.  In initial pilot studies, annotators were asked to choose the ``most intimate questions in each tuple.'' However, this phrasing led to some confusion along two points: (1) lack of an intuitive definition of intimacy that was applicable in many contexts, and (2) how to determine what type of context the question might be asked in. The latter point was important as some questions could be interpreted as more or less intimate when asked in unusual contexts. Based on this feedback, the instructions were revised to (1) describe intimacy with three adjectives,  ``intimate, deep, or personal'' which led to easier judgments, and (2) qualify the question as being asked  ``in the appropriated setting,'' which helped annotators focus less on unusual or abnormal circumstances where a question might be asked. These changes were discussed with pilot annotators and ultimately helped to improve the agreement in further pilot trials. Further, two of the annotators (the authors) were selected to finish all the following annotation tasks. The final annotators first conducted several rounds of training to standardize their judgments and rationale between them. During the training, they independently annotated a small list of tuples and then discussed to resolve the disagreements. After training, the two annotators performed the annotation process as described in the main paper. 

\begin{figure*}[t]
\centering
\includegraphics[width=3.5in]{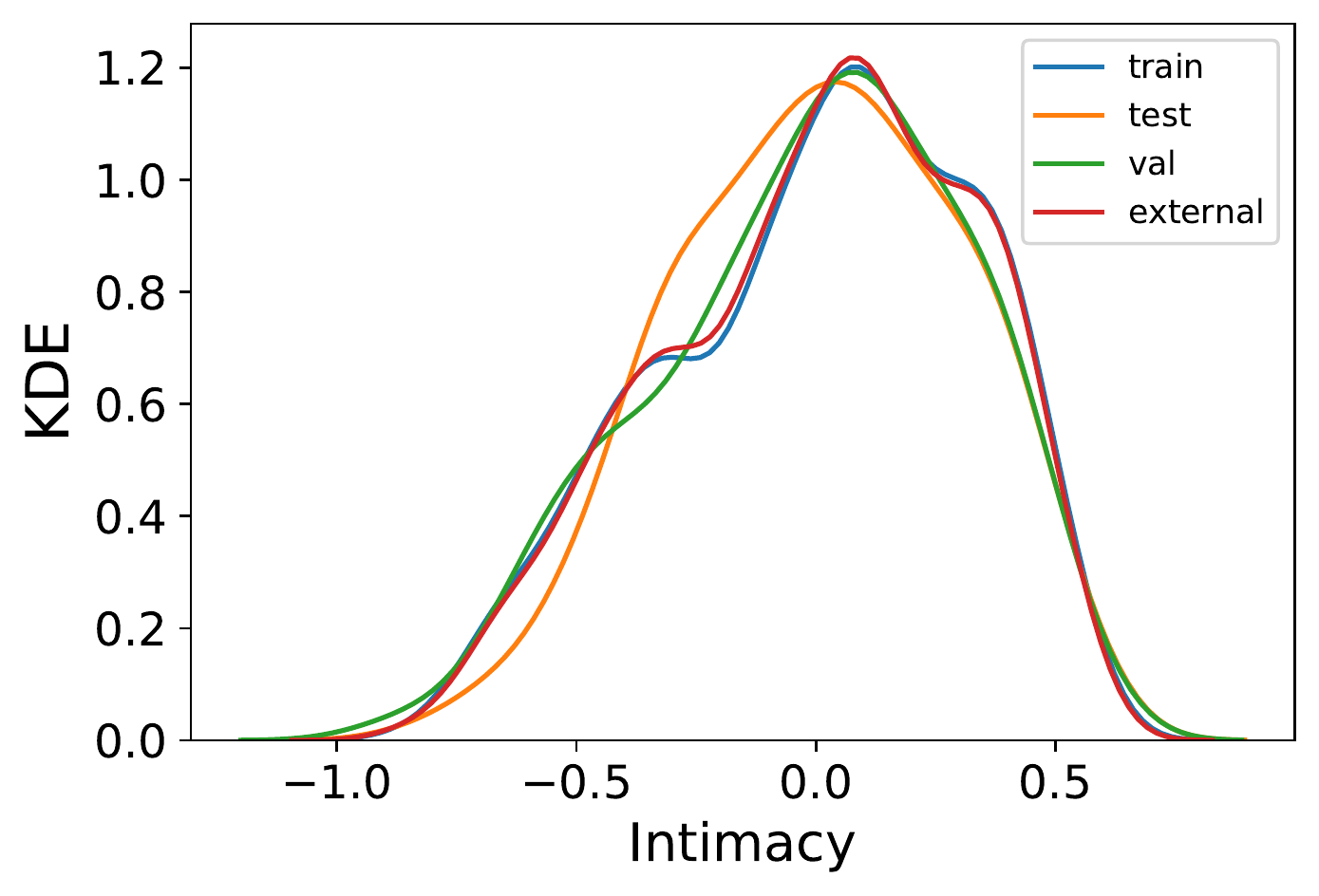}
  \caption{Kernel Density of Annotated Intimacy Dataset. Our dataset is balanced across different splits.}
  \label{fig:data_distplot}
\end{figure*}

\section{Data Samples}
\label{sec:data-sample}
We show the distribution of our annotated question intimacy dataset in Figure \ref{fig:data_distplot} and present data samples in Table \ref{data_sample}. As shown in Figure \ref{fig:data_distplot}, the score distribution across different splits of our dataset is balanced. Further, one can observe the data is slightly skewed to the less intimate (left) side.

\section{Topic Analysis}
\label{sec:topic_analysis}
\begin{figure*}[t]
\centering
\includegraphics[width=.49\linewidth]{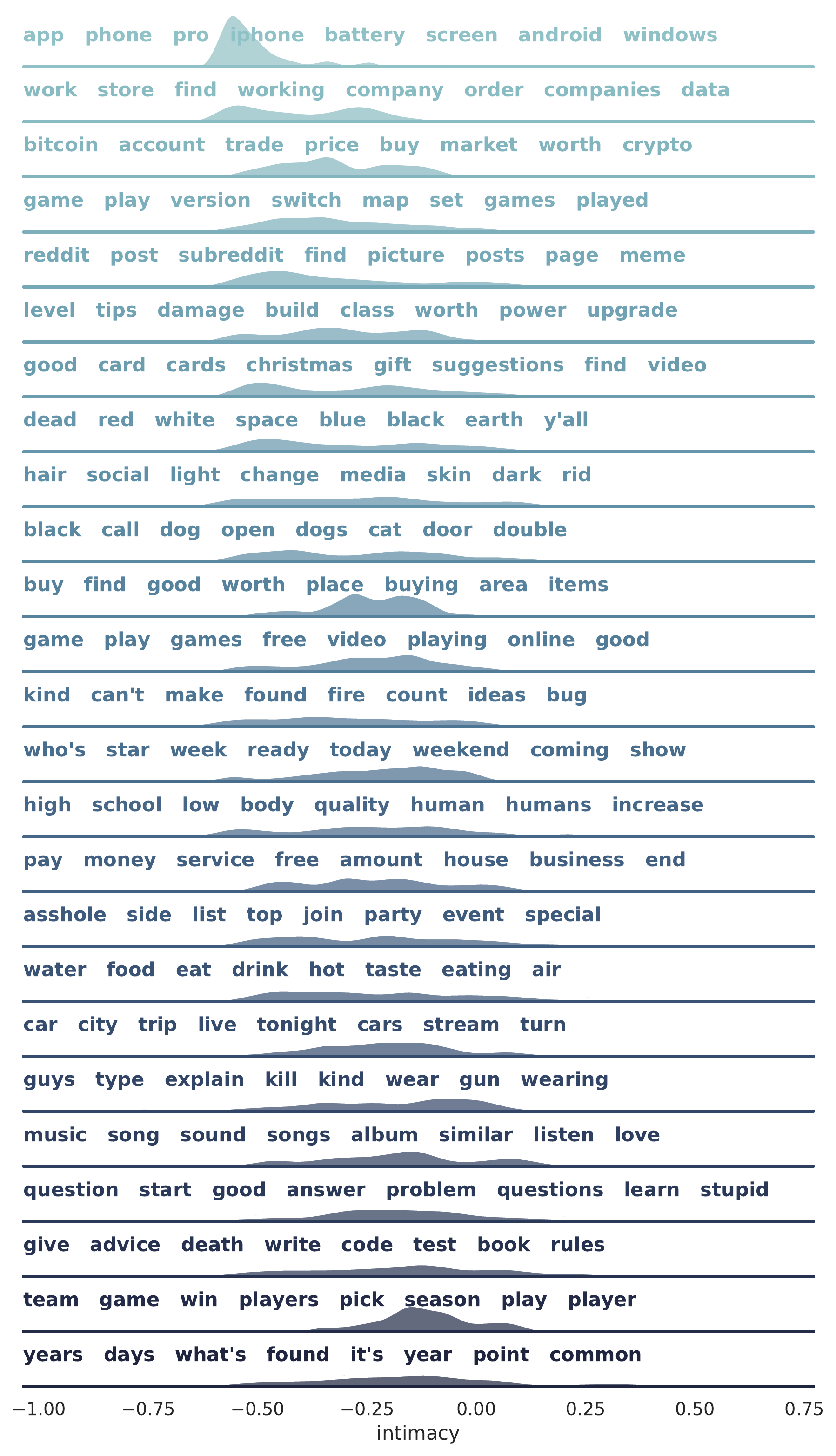}
\includegraphics[width=.49\linewidth]{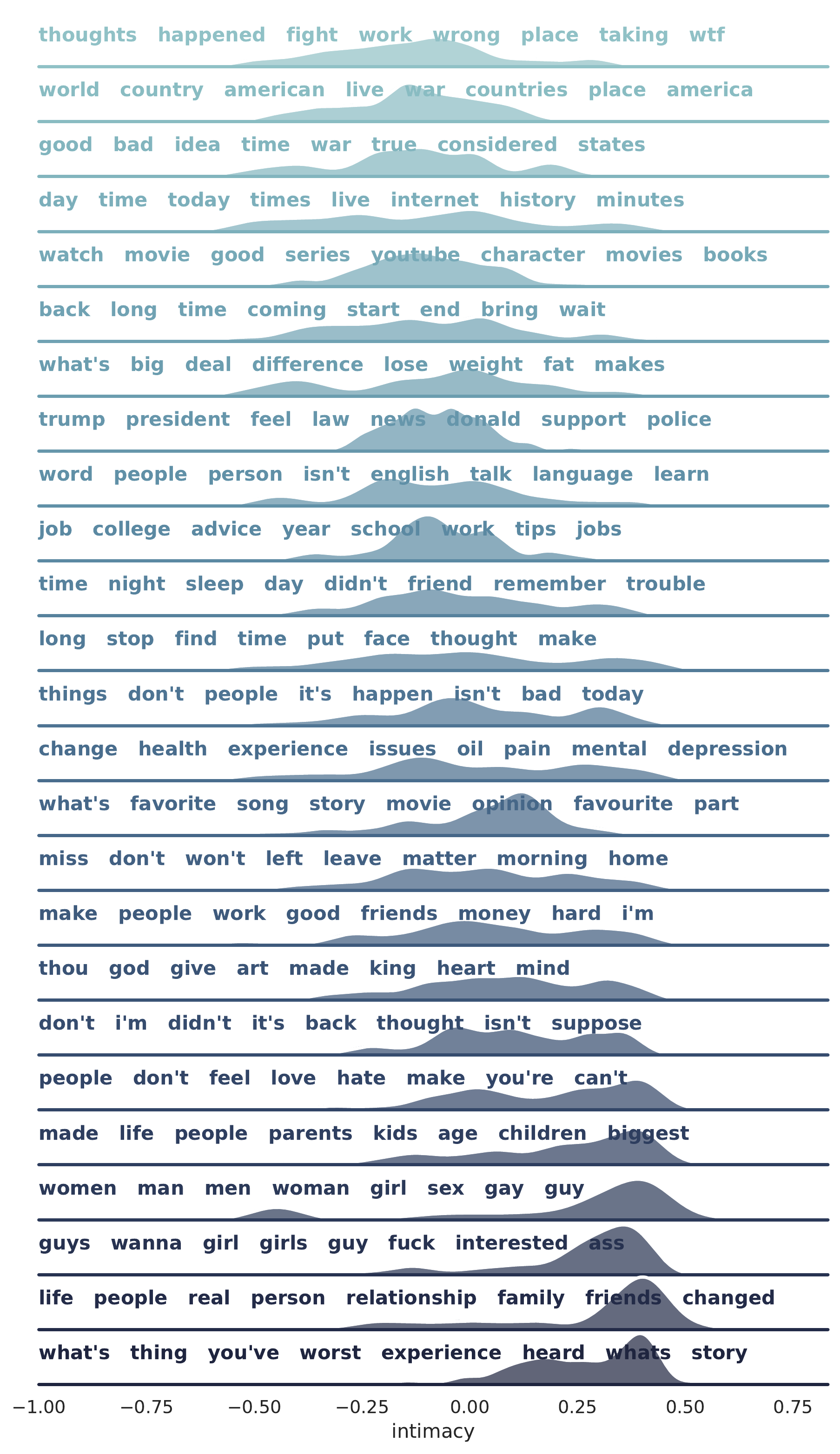}

\caption{Kernel density estimations on the distribution of estimated intimacy scores for the 100 most probable questions for each topic, ordered from least intimate on average (top left) to most intimate (bottom right); these distributions show that many topics exhibit a range of intimacy scores. }
\label{fig:topci_plot}
\end{figure*}

The topic of a question is likely related to the intimacy of a question, with some topics being more taboo and therefore more intimate in nature. To test for this, we trained an LDA topic model using Mallet\footnote{\url{http://mallet.cs.umass.edu/}} to use a question's topic distribution as features for predicting intimacy, as described in Section 4 of the main paper. Here, we report additional experiments on different numbers of topics:  20, 50, 100 and 200 topics. Performance of linear regressors with different numbers of topics are reported in Table \ref{topic_model_comparison}, with the main paper reporting the best-performing of these models. 
Figure \ref{fig:topci_plot} shows the kernel density distribution of intimacy scores for each topic in the 50-topic model, ordered by their mean intimacy. This plot reveals that while some topics are concentrated along specific ranges of intimacy, many span a large range (e.g., finances or weight loss ) and thus topic alone is often insufficient for estimating intimacy. Indeed, even questions with the most intimate topic (on average) that focuses on regretful situations can be asked in less-intimate ways.

\begin{table}[t!]

\newcommand{\tabincell}[2]{\begin{tabular}{@{}#1@{}}#2\end{tabular}}
\centering
\resizebox{0.49\textwidth}{!}{
\begin{tabular}{r cc}

\textbf{Model} &  \textbf{MSE} &  \textbf{Pearson's $r$} \\ 
\hline
LR + 20 topics & 0.06038 & 0.5629 \\
LR + 50 topics  &   \textbf{0.05476} & \textbf{0.6211} \\
LR + 100 topics        & 0.05508 & 0.6055 \\
LR + 200 topics & 0.06136 & 0.5302 \\

\end{tabular}
}
\caption{   Question Intimacy Prediction Performance }
\label{topic_model_comparison}
\end{table}

\section{Pairwise Annotation}
\label{sec:pairwise_annotation}
\begin{figure}[t]
\centering
\includegraphics[width=3.0in]{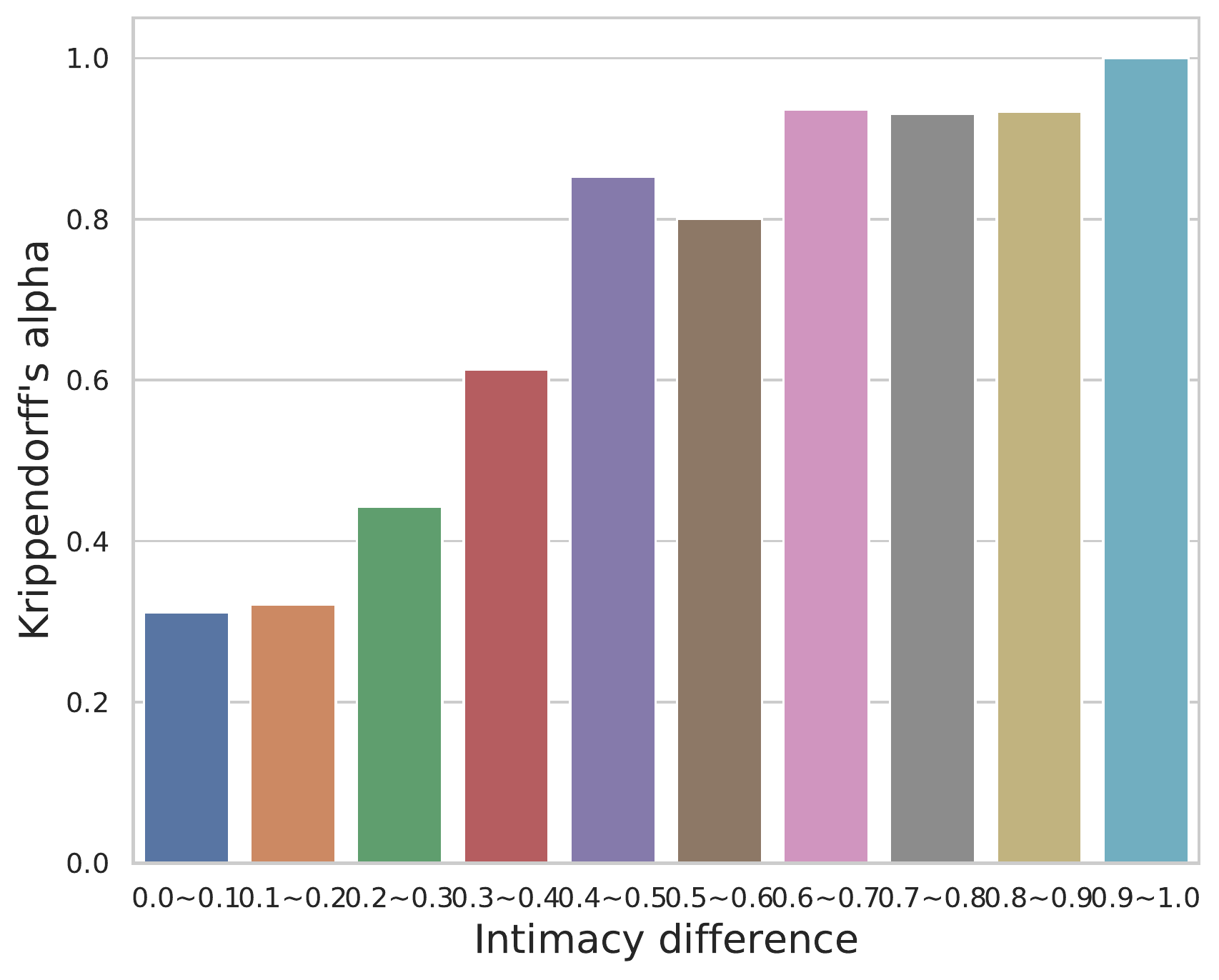}
  \caption{Krippendorff's $\alpha$ at judging which of two questions were more intimate or the same intimacy in the 300 validation annotations sampled from the final dataset. For questions with small model-estimated differences in intimacy, human annotators could not consistently agree on the ranking, resulting in lower $\alpha$; however, the low values are also in part due to the relatively rare frequency of the same-intimacy label, which strongly penalizes $\alpha$. }
  \label{fig:pairwise_iaa}
\end{figure}

\begin{figure}[t]
\centering
\includegraphics[width=3.0in]{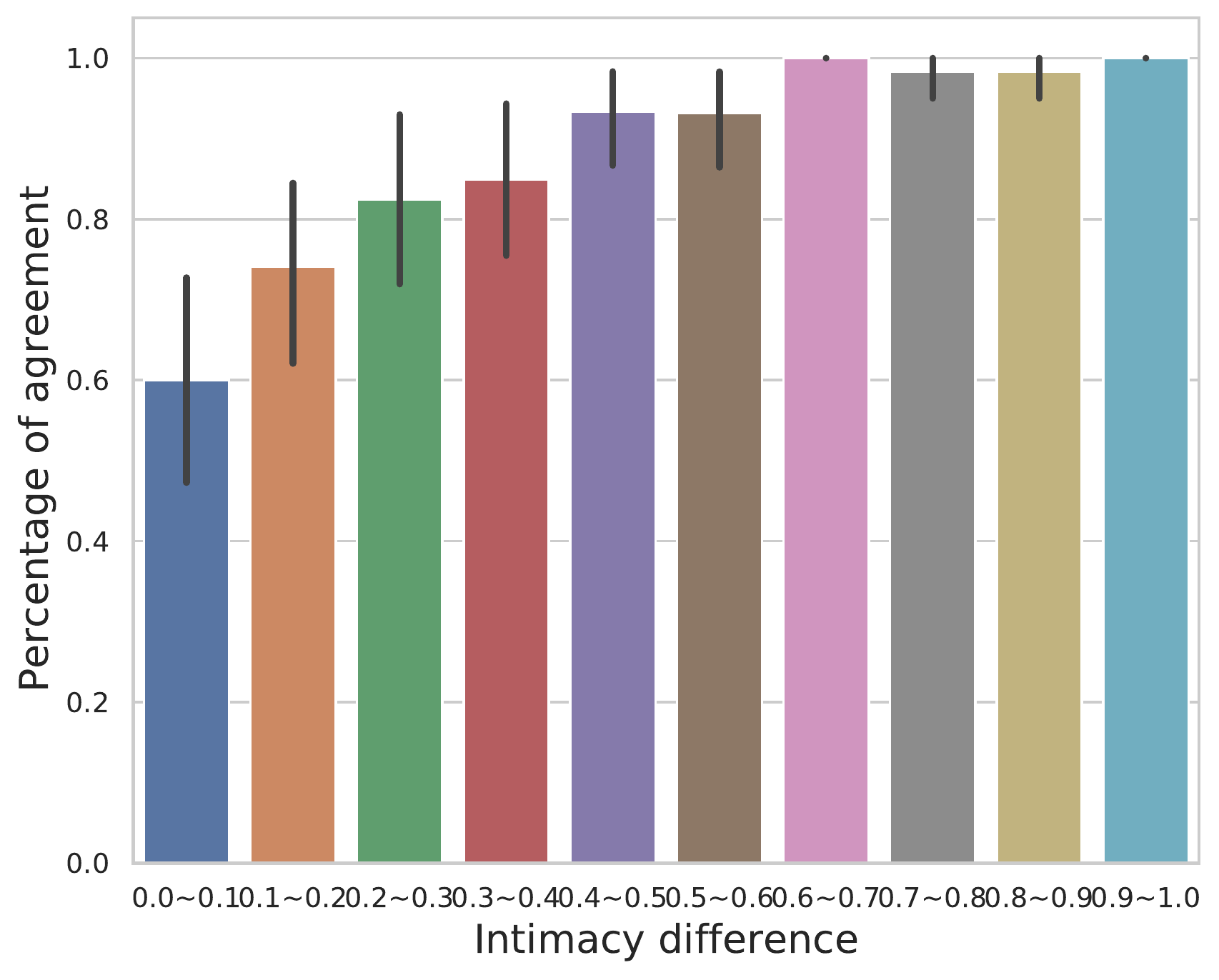}
  \caption{Percentage of agreement between model predictions and human annotations for judging which of two questions were more intimate or the same intimacy in the 300 validation annotations sampled from the final dataset. When the intimacy difference between questions are $\ge$0.2, human annotations are identical to model predictions in over 80\% of the cases. }
  \label{fig:model-agreement}
\end{figure}

As an additional validation on the trained model, both annotators who labeled the initial dataset labeled an additional 300 pairs of questions from the full dataset. Questions were first sorted by their difference in predicted intimacy and binned at 0.1 ranges (e.g., those with distances in $[0.3,0.4)$); then, 30 questions were sampled from each bin to test how sensitive annotators were to each distance. Annotators were asked to select the most intimate of the two question, or if the two questions were too close in similarity to meaningfully describe a difference, to select ``same intimacy.'' Figures \ref{fig:pairwise_iaa} shows Krippendorff's $\alpha$ for the judgments within each bin. Annotators had lower agreement for small differences in intimacy; however, the low values are also in part due to the relatively rare frequency of the same-intimacy label, which strongly penalizes $\alpha$.  Figure \ref{fig:model-agreement} shows the bootstrapped percentage of times the annotators agreed with the models' rank, suggesting that humans largely agree with the model's ranking, especially for large differences in distance.

\begin{table*}[th]
\small

\newcommand{\tabincell}[2]{\begin{tabular}{@{}#1@{}}#2\end{tabular}}
\begin{center}
\begin{tabular}{l|l}
\toprule
\textbf{Question} & \textbf{Intimacy}\\
\midrule
Where is this coin from? & -0.5830 \\
Why is mosquito bites itchy? & -0.5724 \\
Why is the carrying capacity of bags measured in Fluid Ounces? & -0.5472 \\
Why are doctors so afraid of apples? & -0.4906 \\
Best way to stop ears from hurting on a flight? & -0.4876 \\
Would only eating half a vitamin be better? & -0.4616 \\
If everyone in the US switched to using bidets, would the increased water usage be statistically significant? & -0.4522 \\
Does the US shut down for a presidents funeral? & -0.3866 \\
High Liver Enzyme levels? & -0.3590 \\
What are some truly neutral news sources? & -0.3565 \\
Why did Jabba describe the sarlacc as almighty and all-powerful? & -0.3409 \\
Who would become the most dangerous zombie irl? & -0.3070 \\
What is the "John Smith" name of your country? & -0.2508\\
Black Friday shoppers, what do you plan on buying and hopefully getting a deal on? & -0.2435 \\
How did Huey Long's "dictatorship" turn Louisiana from an aristocracy to a true democracy? & -0.2161 \\
Why can't we see that selling weapons for profit is the primary driver of war? & -0.2142 \\
What foods need to be eaten with cutlery and what can be picked up with your hand? & -0.1688 \\
Why do we as a society blame the NRA for gun deaths but not Budweiser etc for drunk driving deaths? & -0.1667 \\
Why was Protestantism adopted in some places and not in others? & -0.1550 \\
What's the most egregious case of a movie trying to make a stand and failing miserably? & -0.1185 \\
Which two solo musical artists would combine to make the best duo? & -0.1164 \\
Why do bloggers, "influencers", etc think they're so important? & -0.0952 \\
What are the easiest laws to accidentally break? & -0.0912 \\
What's the most clever form of cheating you've seen in an exam? & -0.0665 \\
People who forget their phones when they use the bathroom, what do you do? & -0.0533 \\
After having found a tick in my hair, is there a way to get rid of the "creepy crawly" feeling? & -0.0139 \\
What movie would you like to see remade for the special effects? & -0.0136 \\
Is president Donald Trump the second coming of Jesus? & -0.0101 \\
What are your animal stories, wildlife encounters, pet anecdotes, etc? & -0.0082 \\
What are some great tips you learnt from reddit, that you still use today? & -0.0018 \\
Do they like, brush your teeth when you are in coma? & 0.0145 \\
Status of property left behind when moving out of parents house? & 0.0455 \\
If you had infinite resources, how would you improve healthcare? & 0.0754 \\
What do incels do for fun? & 0.0845 \\
What's your favorite thing to cook? & 0.1634 \\
Who got hired without degree, tell us how did you started? & 0.1742 \\
What celebrity(s) do you hate? & 0.2086 \\
Would it be racist for me (a white woman) to dress up as Mulan for a charity event? & 0.2166 \\
What's the craziest thing that ever happened at your school? & 0.2310 \\
What's the weirdest question you have been asked? & 0.3010 \\
How true is it that most girls have fantasies of certain guys graping them? & 0.3012 \\
How do you handle working with an Ex? & 0.3015 \\
Ladys, when a guy your not interested in asks for your number what do you do? & 0.3229 \\
Why is your once best friend not a friend anymore? & 0.3494 \\
What is your story about seeing a dead body outside of relatives in a funeral home? & 0.3610 \\
I've been experiencing intense mood swings and was wondering if this is normal? & 0.3683 \\
What is something so obnoxious makes you sick again and again? & 0.3982 \\
How do religious people know that their god is the "right" god? & 0.4761 \\
\bottomrule

\end{tabular}
\caption{\label{data_sample} Data Samples}
\end{center}
\end{table*}

\end{document}